\documentclass{article}

    \PassOptionsToPackage{numbers, compress}{natbib}

\usepackage{siunitx}
\usepackage[utf8]{inputenc} 
\usepackage[T1]{fontenc}    
\usepackage{hyperref}       
\usepackage{url}            
\usepackage{booktabs}       
\usepackage{amsfonts}       
\usepackage{nicefrac}       
\usepackage{microtype}      
\usepackage{xcolor}         
\usepackage{multirow}

\usepackage{graphicx} 
\usepackage{amsmath} 
\usepackage{fancyhdr} 
\usepackage{geometry} 
\usepackage{caption,subcaption}
\usepackage{tikz,graphics,float,epsf}
\usepackage{xcolor}
\usepackage{cancel}
\usepackage{wrapfig}

\usepackage{hyperref}
\hypersetup{colorlinks, citecolor=blue}

\usepackage[english]{babel}
\usepackage{csquotes}
\usepackage{enumitem}

\usepackage{amsfonts} 
\usepackage{natbib}
\usepackage{algorithm}
\usepackage{algpseudocode}
\usepackage{amssymb}
\usepackage{amsthm}
\usepackage{bbold}
\usepackage{amsmath}
\usepackage{mathtools}
\usepackage{xspace}

\theoremstyle{definition}

\theoremstyle{remark}

\theoremstyle{assumption}

\definecolor{es-blue}{rgb}{0,0.4,0.8}

\title{Maneuver Decision-Making For Autonomous Air Combat Through Curriculum Learning And Reinforcement Learning With Sparse Rewards}

\makeatletter
\newcommand{\printfnsymbol}[1]{%
  \textsuperscript{\@fnsymbol{#1}}%
}

\author{Yu-Jie Wei, Hong-Peng Zhang\thanks{\fontsize{8}{8} 
		Corresponding author.}, Chang-Qiang Huang\\
Aeronautics Engineering College, Air Force Engineering University
}

\begin{document}

\maketitle

\begin{abstract}
Reinforcement learning is an effective way to solve the decision-making problems. It is a meaningful and valuable direction to investigate autonomous air combat maneuver decision-making method based on reinforcement learning. However, when using reinforcement learning to solve the decision-making problems with sparse rewards, such as air combat maneuver decision-making, it costs too much time for training and the performance of the trained agent may not be satisfactory. In order to solve these problems, the method based on curriculum learning is proposed. First, three curricula of air combat maneuver decision-making are designed: angle curriculum, distance curriculum and hybrid curriculum. These courses are used to train air combat agents respectively, and compared with the original method without any curriculum. The training results show that angle curriculum can increase the speed and stability of training, and improve the performance of the agent; distance curriculum can increase the speed and stability of agent training; hybrid curriculum has a negative impact on training, because it makes the agent get stuck at local optimum. The simulation results show that after training, the agent can handle the situations where targets come from different directions, and the maneuver decision results are consistent with the characteristics of missile.

\end{abstract}

\section{Introduction}
\label{sec:intro}

Autonomous air combat maneuver decision-making refers to that the computer alters the control quantities according to the air combat state (such as flight speed, altitude, azimuth, and distance between both sides of air combat) to change the state of motion of the aircraft, so that the aircraft can occupy a valuable position and then attack the target. At present, the research on autonomous air combat maneuver decision-making is still in the stage of simulation, that is, using the kinematics and dynamics model to simulate the flight of the aircraft, and using the maneuver decision-making algorithm to generate the control quantities of the aircraft.

Hu et al.~\citep{hu1} discretized the state space of air combat into a 13 dimensional space for dimension reduction and designed 15 discrete actions to reduce the difficulty of training. With a reward function based on the situation assessment and the final combat gain, the hybrid autonomous maneuver decision strategy was proposed which can realize the capability of obstacle avoidance, formation and confrontation. Dantas et al.~\citep{dan2} compared supervised learning methods using reliable simulated data to evaluate the most effective moment for launching missiles during air combat. They found that the simulated data can improve the flight quality in beyond-visual-range air combat and increase the effectiveness of offensive missions to hit a particular target. Yang et al.~\citep{yang3} constructed a basic maneuver library for the proximal policy optimization algorithm and a reward function with situation reward shaping in order to increase the convergence rate of training. The simulation results shown that the agent with the proposed method can defeat the enemy. Fan et al.~\citep{fan4} proposed a maneuver decision-making method based on asynchronous advantage actor critic algorithm~\citep{hu1}, which incorporates a two-layer reward mechanism of internal reward and sparse reward. This method can reduce the correlation between samples through multi-threading asynchronous learning. Li et al.~\citep{li6} proposed to use the double game tree distributed Monte Carlo search strategy for constructing a two-layer game decision algorithm, and the operational rules of interval numbers and possibility degree comparison rules were adopted to improve the efficiency of strategy searches. Yang et al.~\citep{yang7} proposed an evasive maneuver strategy for beyond-visual-range air combat based on hierarchical multi-objective evolutionary algorithm. They designed four escape maneuvers, including turning maneuver, vertical maneuver, horizontal maneuver and terminal maneuver, and the resigned evolutionary algorithm is used to find approximate Pareto optimal solution and reduce invalid solutions. Wang et al.~\citep{wang8} proposed an autonomous maneuver strategy of aircraft swarms for beyond-visual-range air combat using deep deterministic policy gradient and validated the effectiveness of the method in multi-scene simulations. 

Pope et al.~\citep{pope9} combined a hierarchical architecture with maximum-entropy reinforcement learning, which integrates expert knowledge through reward shaping and supports modularity of policies. The trials shown that the hierarchical architecture achieved human-level performance on air combat. Liu et al.~\citep{liu10} proposed a multi-agent proximal policy optimization algorithm with a comprehensive reward function for an air combat decision-making. The results shown that the algorithm can carry out a multi-aircraft air combat confrontation drill and form new tactical decisions. Zhu et al.~\citep{zhu2021homotopy} proposed a homotopy-based soft actor-critic method to address the sparse reward problem, which follows the homotopy path between the original task with sparse reward and the auxiliary task with artificial prior experience reward. Experimental results shown that the method performed better than that utilizing the sparse reward or the artificial prior experience reward. Ozbek et al.~\citep{ozbek2022reinforcement} used twin delayed deep deterministic policy gradients and hindsight experience replay to find an optimal path to approach the target in two dimensional space. Wang et al.~\citep{yuan2022research} focused on the problem of insufficient exploration ability of Ornstein-Uhlenbeck strategy and proposed a heuristic algorithm with heuristic exploration strategy. To improve the maneuver decision ability of manual manipulation for autonomous air combat, Li et al.~\citep{li2022autonomous} proposed to combine the simulated operation command with the final reward value deep deterministic policy gradient algorithm and designed a prediction method to simulate the operation command of the enemy aircraft. The results shown that the algorithm can improve the maneuver decision-making ability. Proximal policy optimization (PPO)~\citep{engstrom2020implementation} is used to train the air combat agent in  

Air combat maneuver decision-making with sparse reward is an interesting but difficult problem. In air combat maneuver decision-making, sparse rewards mean that the agent can acquire the rewards only when the air combat ends, and the agent cannot acquire any handcrafted rewards before the ending of the air combat. Since the reward signals are sparse, it is difficult for the agent to obtain effective training samples, which may lead to poor training performance. On the other hand, if we can train the agent successfully by means of sparse rewards, it means that the agent can make maneuver decisions without any human knowledge~\citep{chen2009monte,silver2017mastering,fawzi2022discovering}, which is an interesting direction of air combat maneuver decision-making. Meanwhile, using sparse rewards can avoid designing reward functions, which is a time-consuming and laborious procedure. Different from the above literatures, the goal of this paper is to train the agent by means of sparse reward, that is, without using any handcrafted reward function, and to enable the trained agent to cope with targets from different directions (that is, the azimuth range of the target is $[-\pi,\pi]$).

However, when applying reinforcement learning to maneuver decision-making, using only sparse rewards may result in too much training time, or even training failure. Therefore, in order to acquire the effective air combat agent and reduce training time as much as possible, we propose to integrate curriculum learning into reinforcement learning. The main contributions of this article are as follows: 1. According to the characteristics of air combat, three different curricula are designed to train agents, namely, angle curriculum, distance curriculum and hybrid curriculum. 2. In the framework of self-play,proximal policy optimization (PPO)~\citep{engstrom2020implementation} is combined with the above three curricula, and only sparse rewards are used to train agents. 3. The ablation experiments are conducted to investigate the characteristics of the three curricula. 4. The simulation experiments are conducted to verify the effectiveness of the agent.

\section{Related work}
\label{sec:pre}
Elman pointed out that successful learning may depend on starting from small things~\citep{elman1993learning}. To test the neural network's ability to learn and express the relationship between parts and the whole, Elman trained the neural network to process complex sentences. These networks can learn to solve the task only when they are forced to start from strict memory restrictions, which is equivalent to limiting the range of data provided to the network at the initial learning stage. Bengio formalized such training strategies in machine learning and named them as currirulum learning~\citep{bengio2009curriculum}. For example, when people and animals are provided with examples in a meaningful order (such as gradually increasing the difficulty or quantity) rather than random order, they can learn better. 

Jiang et al. proposed a self-paced curriculum learning~\citep{jiang2015self}, which takes into account both the prior knowledge before the training and the learning process during the training. Based on the generalized boundary criterion of the task order~\citep{pentina2015curriculum}, Pentina et al. optimized the average expected classification performance on all tasks, and solved the problem of multiple task curriculum learning by finding the best task order. Xing et al. sorted the samples according to the complexities~\citep{sachan2016easy}, so that the simpler samples can be used in the learning algorithm earlier, and the more difficult samples can be used later. They proposed seven heuristic methods to improve curriculum learning, and compared these methods on four non-convex question answering models. The experimental results show that these methods can improve the performance. Graves et al. introduced a method of automatically selecting curricula according to the growth of prediction accuracy and the growth of the network complexity to maximize the learning efficiency~\citep{graves2017automated}. They used the proposed method to train LSTM networks~\citep{hochreiter1997long} and the experimental results show that the method can significantly accelerate the learning speed. Jiang et al. proposed a new curriculum learning method called MentorNet~\citep{jiang2018mentornet}. The method dynamically learns a data-driven curriculum to overcome the overfitting problem of damaged labels. The experimental results show that the method can improve the generalization performance of deep networks trained on corrupted data. Zhou et al. proposed a minimax curriculum learning (MCL) to adaptively select subset sequences of training in a series of stages of machine learning~\citep{zhou2018minimax}. The results show that MCL achieves better performance and uses fewer samples for training both shallow and deep models while achieving the same performance. Platanios et al. proposed a curriculum learning framework for neural machine translation~\citep{platanios2019competence}, which determines the training samples displayed to the model at different times during training according to the estimated difficulty of the samples and the current ability of the model. The experimental results show that the proposed framework can reduce training time, reduce the demand for professional heuristic methods and large quantities of data, and make the overall performance better.

Inspired by human learning, Stretcu et al. proposed a novel curriculum learning method, which decomposes challenging tasks into easier intermediate sequences for pre-training the model before processing the original tasks~\citep{stretcu2020coarse,stretcu2021coarse}. They trained the model at each level of the hierarchy from coarse labels to fine labels, so as to transfer the acquired knowledge at these levels. The results show that the classification accuracy of the method is improved by 7\%. Zhao et al. put forward a formulation of multitask learning~\citep{zhao2020efficient}. The formulation learns the relationship between tasks represented by a task covariance matrix and the relationship between features represented by a feature covariance matrix. Li et al. proposed a competence-aware curriculum for visual concept learning by question and answer manner~\citep{li2020competence}. The method includes a neural symbol concept learner for learning visual concepts and a multi-dimensional Item Response Theory model for guiding the learning process with adaptive curriculum. The experimental results show that the proposed method achieves the most advanced performance with excellent data efficiency and convergence speed through the competence-aware curriculum. Wu et al. studied the impact of limited training time budget and noisy data on curriculum learning~\citep{wu2020curricula}. The experimental results show that, under the condition of limited training time budget or noise data, curriculum learning rather than anti curriculum learning can indeed improve the learning performance. In order to improve the ability of convolutional neural network (CNN)~\citep{lecun2015deep,lawrence1997face,turaga2010convolutional} of representing shape and texture information at the same time, Sinha et al. proposed to gradually increase the amount of texture information available in training by reducing the standard deviation of the Gaussian kernel~\citep{sinha2020curriculum}. This training scheme significantly improves the performance of CNN in various image classification tasks without adding additional trainable parameters or auxiliary regularization targets. Weinshall et al. provided theoretical research on curriculum learning when using stochastic gradient descent to optimize convex linear regression loss~\citep{weinshall2018curriculum}, and proved that the convergence rate of the ideal curriculum learning method monotonously increased with the difficulty of the examples. In order to analyze the impact of curriculum learning on the training of deep convolution network for image recognition, Hacohen et al. proposed two methods~\citep{hacohen2019power}, namely, transfer learning and bootstrapping, to sort training examples by difficulty, and used different pace functions to guide the sampling.

In addition to solving problems in deep learning, curriculum learning is also used to solve problems in reinforcement learning~\citep{williams1992simple,bertsekas2019reinforcement}. Fournier et al. proposed a curriculum learning method based on accuracy~\citep{fournier2018accuracy}. The method is based on deep deterministic policy gradient algorithm~\citep{shi2018multi,barth2018distributed}, adaptively select the requirements of accuracy, and automatically generate curricula with increasing difficulty. The results show that the method can improve the learning efficiency. Narvekar et al. proposed a method to automatically generate task sequences for special curricula in reinforcement learning~\citep{narvekar2017autonomous}. The method uses the heuristic function in~\citep{narvekar2016source} to recursively decompose difficult tasks, and selects tasks that results in the greatest change in policy. Florensa et al. proposed an reverse curriculum generation method to solve the problem of reinforcement learning~\citep{florensa2017reverse}. The method uses "reverse learning" to solve difficult robot operation tasks without any prior knowledge. The robot is trained to reach a given target state from a nearby initial state at first. Then, the robot is trained to solve the task from a further initial state. Rane uses curriculum learning to solve tasks with sparse rewards in deep reinforcement learning~\citep{rane2020learning}. The experimental results show that curriculum learning can improve the convergence speed of training and make the performance better.

\section{Method}
\label{sec:method}

\subsection{Models} 
The kinematic and dynamic model of the aircraft is shown as follows~\citep{hu1}:
\begin{align}
\label{eq:minimax}
    \begin{cases}
    \dot{x}=v\cos\gamma\cos\phi\\
    \dot{y}=v\cos\gamma\sin\phi\\
    \dot{z}=v\sin\gamma\\
    \dot{v}=g(n_x-\sin\gamma)\\
    \dot{\gamma}=\frac{g}{v}(n_z\cos\mu-\cos\gamma)\\
    \dot{\psi}=\frac{g}{v\cos\gamma}n_z\sin\mu\\
    \end{cases}
\end{align}

When the minimum distance between the missile and the target is less than 12 m, the target is regarded as being hit; when the missile flight time exceeds 120 s, and it still fails to hit the target, the target is regarded as missed; during the midcourse guidance stage, the target is regarded as missed when its azimuth relative to the aircraft exceeds $\pi/3$; during the final guidance stage, the target is regarded as missed when its azimuth relative to missile axis exceeds $\pi/2$. The conditions for the end of air combat simulation are: 1. One of the both side in air combat is hit by the missile; 2. The missiles of the both sides miss the targets; 3. The maximum time of simulation is reached, which is set as 200 s in this article. Currently, most of the air combat maneuver decision-making methods use handcrafted reward functions, but designing reward functions requires plenty of time and related knowledge. Moreover, inappropriate reward functions may cause unexpected or unreachable behaviors. Therefore, we do not use any handcrafted reward function. Namely, the reward obtained by the agent is 1 when it defeats the opponent, $-1$  when it is defeated by the opponent, and 0 in other cases. 

\subsection{Curriculum learning} 

Curriculum learning~\citep{elman1993learning} is inspired by the process of human learning in the real world: Humans do not learn difficult tasks from scratch, on the contrary, they start from simpler tasks and gradually learn to solve more difficult tasks. For example, a student begins to learn addition, subtraction, multiplication and division, function and limit at first, then, he tries to learn differential and integral calculus. Inspired by human learning, curriculum learning is a learning strategy for machine learning, that is, first learn to solve simple tasks, then improve the difficulty of tasks and learn to solve these more difficult tasks.

We combine the characteristics of air combat with curriculum learning and propose three different curricula to investigate the impact of different curricula on the training process and the performance of agents, and to improve the ability of air combat maneuver decision-making. The agent training strategies based on course learning are as follows: 1. Curriculum design. First, design the difficulties of the curricula, and determine the condition for curriculum transfer (namely, increasing the curriculum difficulty). 2. Curriculum learning. Train the agent in the initial curriculum. After the training, test the the agent. If the test results meet the condition for curriculum transfer, the agent can be trained in the next curriculum, otherwise, in the current curriculum. The three different curricula are shown below.

\subsubsection{Angle curriculum}

In the air combat, the larger azimuth of the target means the more difficult it is for the agent to hit the target. First, the agent can launch the missile without any maneuver and as long as the target is within the detection range of the radar, the missile may not miss the target in the midcourse guidance phase. In addition, the smaller azimuth means that it takes more time for the target to escape from the detection of the radar. In conclusion, the smaller azimuth of the target, the less difficult it is for the agent to defeat the target; the larger azimuth of the target, the more difficult it is for the agent to defeat the target. Therefore, the angle curriculum is designed according to the initial azimuth of the target, which varies in: $\left[\frac{-\pi}{10},\frac{\pi}{10}\right]$, 
$\left[\frac{-\pi}{5},\frac{\pi}{5}\right]$, ...,
 $\left[-\pi,\pi\right]$. Meanwhile, the initial distance is randomly and uniformly sampled from the interval of $\left[50,000~\si{\metre}, 150,000~\si{\metre}\right]$.

\subsubsection{Distance curriculum}

The smaller the distance between the two sides, the less time it takes for the midcourse guidance of the missile, thus, the less difficult it is for the agent to defeat the target. Therefore, distance curriculum is designed according to the initial distance between the two sides of air combat, which varies in: $\left[50,000~\si{\metre}, 60,000~\si{\metre}\right]$, $\left[50,000~\si{\metre}, 70,000~\si{\metre}\right]$, ..., $\left[50,000~\si{\metre}, 150,000~\si{\metre}\right]$, that is, the distance range of the next curriculum is 10,000 m larger than that of the previous curriculum. Meanwhile, the initial azimuth of the target is randomly and uniformly sampled from the interval of $\left[-\pi,\pi\right]$.

\subsubsection{Hybrid curriculum}

The hybrid curriculum combines the angle curriculum with the distance curriculum. In hybrid curriculum, the initial azimuth of the target and the initial distance between the both sides are increased simultaneously. Namely, the angle range of the next curriculum is $\frac{\pi}{5}$ greater than that of the previous curriculum, and the distance range of the next curriculum is 10,000~\si{\metre} greater than that of the previous curriculum.

\section{Experiments} 
\label{sec:exp}
In order to verify the performance of the training results of the three different curricula and the performance of the trained agents, ablation studies and simulations are performed in this section. For the three different curricula and the original method without curriculum, we perform five times of independent training. Each training consists of forty iterations, and each iteration consists of twenty times of the cycle of Figure 5. Table~\ref{tab:hyper} lists the hyperparameters.

\begin{table}[]
	\centering
	\caption{Hyperparameters}
	\label{tab:hyper}
	\begin{tabular}{|c|c|}
		\hline
		\textbf{Name  }     & \textbf{Value}  \\ \hline
		Velocity  & $\left[250~\si{\meter\per\second},400~\si{\meter\per\second}\right]$   \\ \hline
		Batchsize                 & 1024   \\ \hline
		Optimizer                 & Adam  \\ \hline
		Actor learning rate                 & 0.002  \\ \hline
		Critic learning rate                 & 0.001  \\ \hline
		Actor architecture            & 256*256*4  \\ \hline
		Critic architecture         & 256*256*1  \\ \hline
		Activate function                 & tanh  \\ \hline
		Epoach                 & 8  \\ \hline
		$\gamma$                           & 1  \\ \hline
	\end{tabular}
\end{table}

\begin{figure}[]
	\vspace{-9ex}
	\centering
	\begin{subfigure}[b]{0.5\textwidth}
		\centering
		\includegraphics[width=\textwidth]{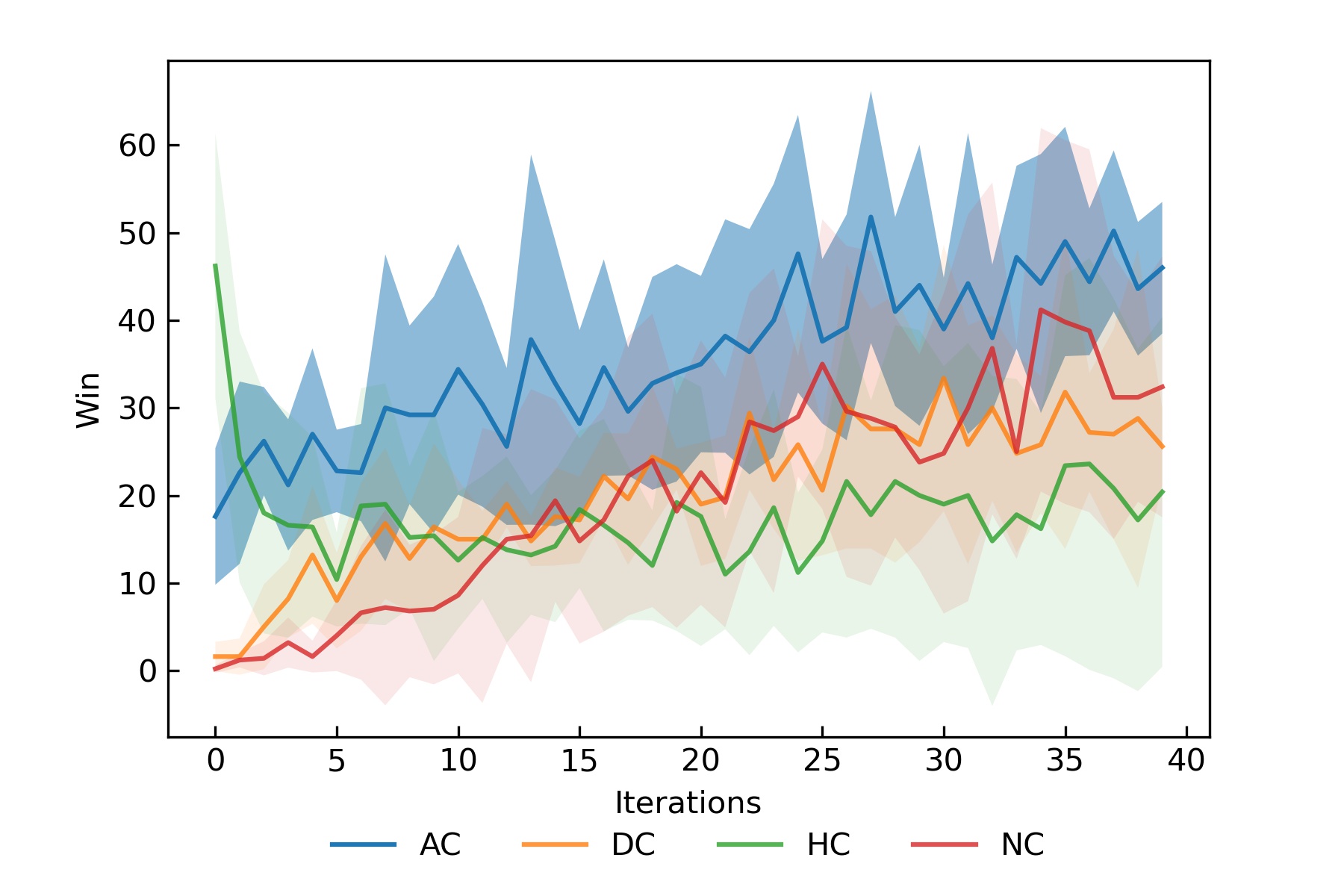}
		\caption{}
		\label{fig:win}
	\end{subfigure}
	\vfill
	\begin{subfigure}[b]{0.5\textwidth}
		\centering
		\includegraphics[width=\textwidth]{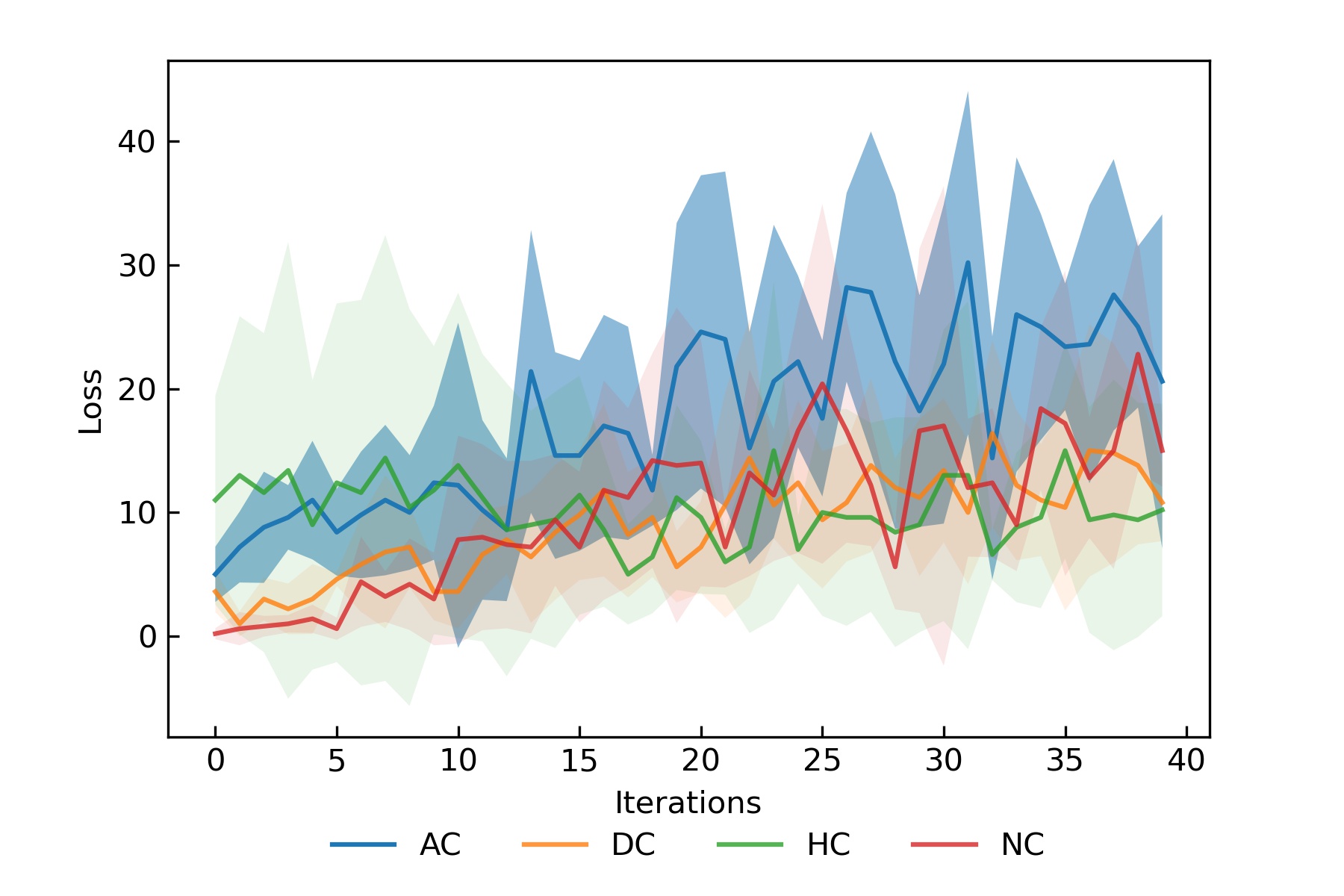}
		\caption{}
		\label{fig:loss}
	\end{subfigure}
	\vfill
	\begin{subfigure}[b]{0.5\textwidth}
		\centering
		\includegraphics[width=\textwidth]{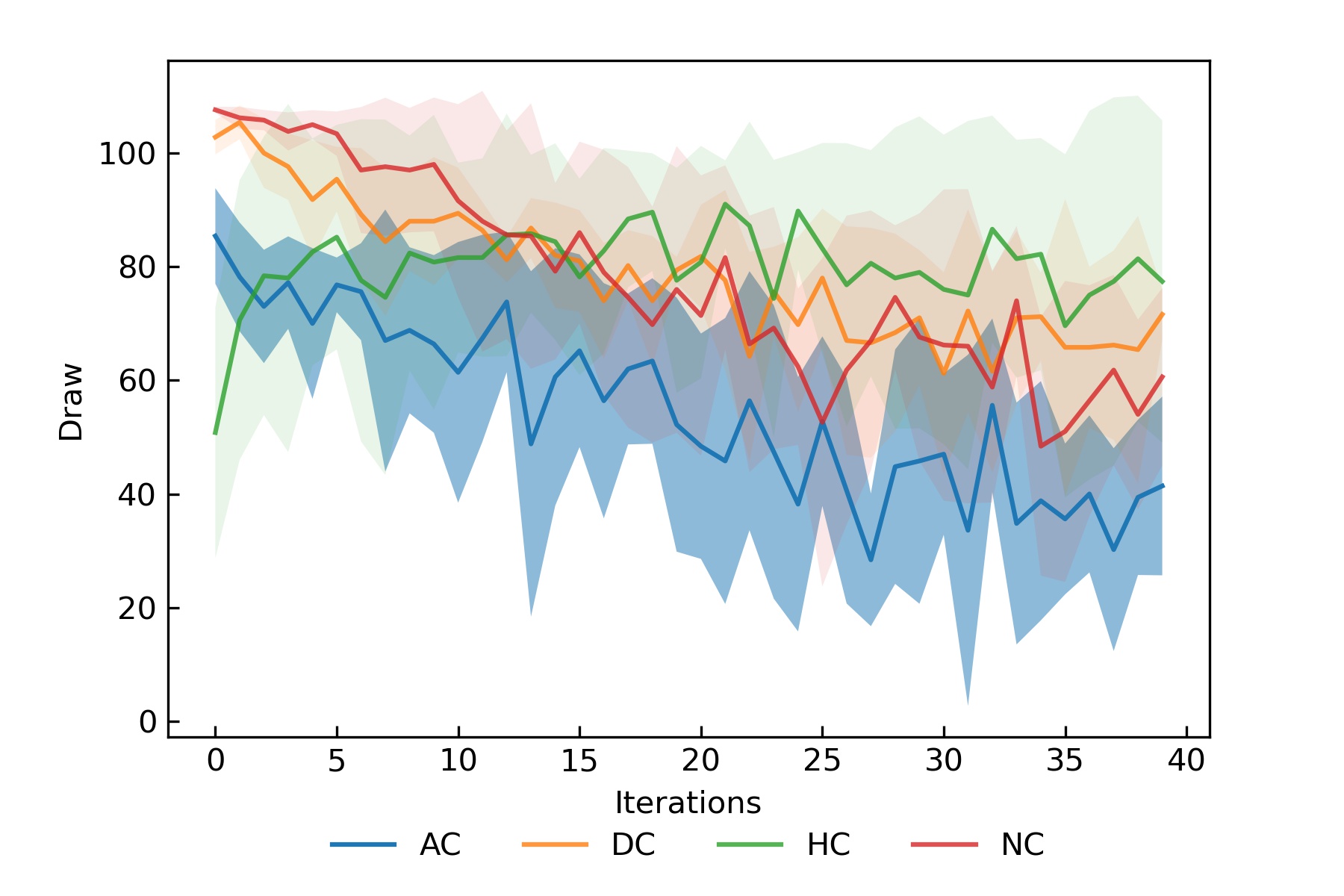}
		\caption{}
		\label{fig:draw}
	\end{subfigure}
	\vfill
	\caption{\textbf{(a)} Win. \textbf{(b)} Loss. \textbf{(c)} Draw.}
	\vspace{-4ex}
	\label{fig:train}
\end{figure}

In this section, the processes of training agents using four different methods are compared: angle curriculum (AC), distance curriculum (DC), hybrid curriculum (HC), and no curriculum (NC), the original method without any curriculum). The training processes of the four methods at each iteration can be seen in Figure~\ref{fig:train}. The solid line represents the mean of the number of win, loss or draw of the corresponding curriculum, and the shaded part represents the standard deviation of the number of win, loss or draw. 

As shown in Figure~\ref{fig:train}, some shaded parts are negative. However, this phenomenon does not mean that the number of win or loss are negative. It mainly because that the mean values are less than the corresponding standard deviations, which results in the negative shaded parts. Obviously, the more win and the fewer draw represent the better training performance. At the same time, more loss can also indicate that the training performance is better, because both the win and the loss are the results of the agent itself.

As shown in Figure~\ref{fig:win}, during the training, the wins of angle curriculum, distance curriculum and no curriculum are gradually increasing, and the win of angle curriculum is always the most. The win of distance curriculum rises faster than that of no curriculum, and the standard deviation of distance curriculum is less. At the beginning of the training, the win of hybrid curriculum is the most, and then it suddenly decreases and remains almost unchanged. It can be seen from Figure~\ref{fig:loss} that the loss of angle curriculum, distance curriculum and no curriculum are gradually increasing, while the loss of hybrid curriculum has no obvious trend of increasing or decreasing. It can be seen from Figure~\ref{fig:draw} that during the training, the draws of angle curriculum, distance curriculum and no curriculum are decreasing, and the draw of angle curriculum is less. The draw of hybrid curriculum is the least at the beginning of the training, and then it increases and remains almost unchanged.

\section{Discussion}
\label{sec:Discussion}
According to the ablation studies in Section~\ref{sec:exp}, angle curriculum is the best, hybrid curriculum is the worst, and distance curriculum is better than hybrid curriculum but worse than angle curriculum. Although the number of win of no curriculum is slightly more than that of distance curricula in the late stage of training, the training of it is slower and more unstable. However, angle curriculum can not only accelerate the training, but also significantly improve the number of win. Hybrid curriculum is useless, which is mainly because that the agent gets stuck at local optimum in the initial curriculum learning, resulting in the failure of curriculum transfer.
Therefore, in curriculum learning, the curriculum design depends on professional knowledge and common sense of the certain fields. The designed curriculum may not only improve the training speed and performance, but also cause overfitting and failure. For example, although the hybrid curriculum is consistent with the common sense of the air combat, it is invalid. Overall, appropriate curricula can accelerate the training and provide better performance while improper curricula can damage the training.

\section{Conclusion}
\label{sec:Conclusion}
The purpose of this paper is to propose an effective reinforcement learning method for air combat maneuver decision-making with sparse rewards. Although sparse rewards is not limited by human experience, getting reward signals is hard for the agent, which results in much training time and unstable training process, and even failure of accomplishing the task.
In order to solve these problems, the method based on curriculum learning and reinforcement learning is proposed. First, three curricula are designed: angle curriculum, distance curriculum and hybrid curriculum. Then, the curricula are used to train agents for air combat maneuver decision-making. The training results indicate that the performance of angle curriculum is the best, which can not only improve the speed and stability of training, but also improve the performance of the agent; distance curriculum can improve the speed and stability of training; hybrid curriculum is invalid, because it makes the agent get stuck at local optimum at the initial stage of training, which leads to the failure of curriculum transfer. More importantly, sparse rewards is enough for effective air combat agents and the handcrafted reward functions are not necessary. 

\subsubsection*{Acknowledgments}
\vspace{-0.5em}
\small{
We thank Ali Wang for various help and feedback.}

\normalsize
\bibliography{main.bib}

\end{document}